\documentclass{article}

\usepackage{enumitem}

\usepackage{float} 
\usepackage[preprint]{neurips_2025}
\usepackage{comment}
\usepackage{algorithm}   
\usepackage{tabularx}
\usepackage{algpseudocode}        
\usepackage{booktabs}
\usepackage{natbib}

\usepackage{subcaption}   

\usepackage{wrapfig}
\usepackage{array}
\newcolumntype{C}[1]{>{\centering\arraybackslash}p{#1}}
\usepackage[utf8]{inputenc} 
\usepackage[T1]{fontenc}    
\usepackage{hyperref}       
\usepackage{url}            
\usepackage{booktabs}       
\usepackage{amsfonts}       
\usepackage{nicefrac}       
\usepackage{microtype}      
\usepackage{xcolor}         
\usepackage{amsmath}
\usepackage{graphicx}
\usepackage[compact]{titlesec}
\titlespacing{\section}{0.25pt}{*0}{*0}
\titlespacing{\subsection}{0.1pt}{*0}{*0}
\titlespacing{\subsubsection}{1pt}{*0}{*0}
\title{Text embedding models can be great data engineers}

\author{%
  Iman Kazemian$^{1}$, \quad
  Paritosh Ramanan$^{2}$, \quad
  Murat Yildirim$^{1}$\thanks{Corresponding author: murat@wayne.edu} \\
  $^{1}$Wayne State University \quad
  $^{2}$Oklahoma State University
}

\begin{document}

\maketitle
\vspace{-0.5cm}

\begin{abstract}

Data engineering pipelines are essential - \textit{albeit costly} - components of predictive analytics frameworks requiring significant engineering time and domain expertise for carrying out tasks such as data ingestion, preprocessing, feature extraction, and feature engineering. In this paper, we propose \textit{ADEPT}, an \underline{a}utomated \underline{d}ata \underline{e}ngineering \underline{p}ipeline via \underline{t}ext embeddings. At the core of the \textit{ADEPT} framework is a simple yet powerful idea that the entropy of embeddings corresponding to textually dense raw format representation of timeseries can be intuitively viewed as equivalent \textit{(or in many cases superior)} to that of numerically dense vector representations obtained by data engineering pipelines. Consequently, \textit{ADEPT} uses a two step approach that (i) leverages text embeddings to represent the diverse data sources, and (ii) constructs a variational information bottleneck criteria to mitigate entropy variance in text embeddings of time series data. ADEPT provides an end-to-end automated implementation of predictive models that offers superior predictive performance despite issues such as missing data, ill-formed records, improper or corrupted data formats and irregular timestamps. Through exhaustive experiments, we show that the \textit{ADEPT} outperforms the best existing benchmarks in a diverse set of datasets from large-scale applications across healthcare, finance, science and industrial internet of things. Our results show that \textit{ADEPT} can potentially leapfrog many conventional data pipeline steps thereby paving the way for efficient and scalable automation pathways for diverse data science applications. 

\end{abstract}

\section{Introduction}
Data engineering pipelines are fundamental components for enabling predictive analytics on time series data in several areas such as energy \cite{rahimilarki2022convolutional}, healthcare \cite{an2023comprehensive} and finance \cite{dingli2017financial}. These pipelines broadly comprise of seven sequential steps pertaining to data ingestion; data preprocessing; feature extraction; feature engineering; model training and testing followed by model deployment \cite{raj2020modelling}. Preprocessing typically involves data cleaning mechanisms that aim to eliminate ill-formed records \cite{felix2019systematic}, resolve irregularities in sampling as well as impute missing values. Feature engineering and extraction steps deal with identifying features of the input timeseries encoding the information most relevant to the predictive analytics task at hand \cite{lin2020missing}. For best efficiency gains, it becomes necessary to carefully customize methodological frameworks used for data cleaning, feature engineering and extraction tasks with respect to the application specific domain area and data challenges. As a result, despite several advances in the field of autoML \cite{salehin2024automl}, automating the data cleaning, feature engineering and extraction steps remain one of the most challenging and expensive tasks across conventional data engineering pipelines due to the need for significant manual intervention  and domain expertise \cite{salehin2024automl}. In this paper, we present ADEPT, a framework that attempts to drastically simplify data engineering pipeline complexity by applying LLM-based text embedding models on raw text representations of input timeseries as a precursor to the model training step. 

Fundamentally, the pattern recognition capability of any predictive model is a direct consequence of capturing temporal and spatial correlations in the timeseries input. From an information theoretic perspective, we argue that the entropy of embeddings corresponding to textually dense raw format representation (RFR) of timeseries (such as CSV, HDF5 etc.) can be intuitively viewed as equivalent to that of numerically dense vector representations obtained by data engineering pipelines. As a result, LLM-based text embeddings of timeseries RFRs can \emph{also potentially be seen as alternative representations of spatial and temporal correlations essential for training a predictive model}. Consequently, ADEPT enables a significantly simpler data representation that can be used for model training while retaining its spatiotemporal aspects. 
Also, the \textit{ADEPT} framework exploits text embeddings of timeseries RFRs to effectively leapfrog data cleaning, feature engineering and extraction steps of data engineering pipelines. In doing so, ADEPT demonstrates significant resiliency with respect to missing data, ill-formed records, improper or corrupted data formats as well as irregular timestamps. 


\begin{wrapfigure}{r}
  {0.7\textwidth}
  \vspace{-5mm}
  \centering
\includegraphics[width=0.7\textwidth]{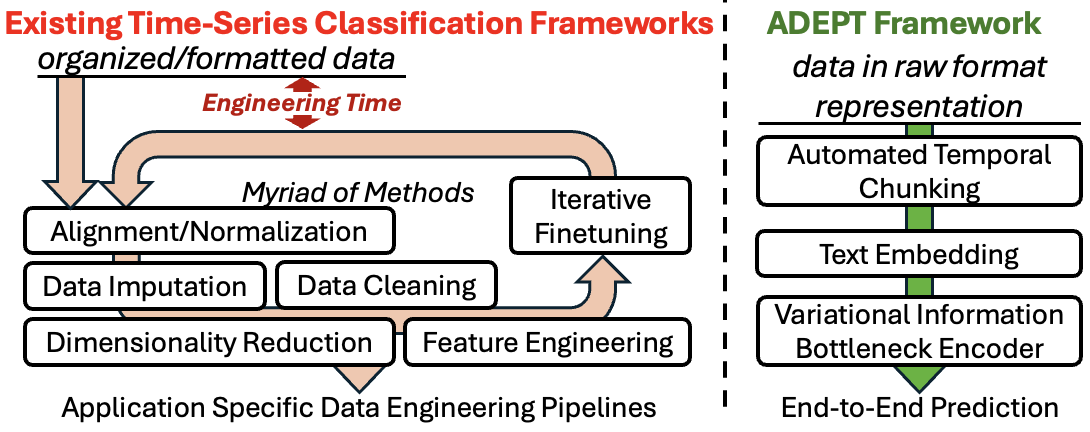}
  \caption{Comparison of the model and benchmark.}
  \label{fig:overview}
  \vspace{-4mm}
\end{wrapfigure}
The methodological contribution of ADEPT relies on exploiting text embedding models primarily geared for LLM use cases as the foundational building block to power timeseries oriented predictive analytics tasks. As a result, the ADEPT framework leverages LLM based text embedding models as a \emph{black box}, eliminating need for complex fine-tuning and retraining tasks. We introduce a Variational Information Bottleneck (VIB) criteria as a means to reduce the entropy variance emanating from text embeddings of timeseries RFR. Additionally, the VIB criteria is used to train a multi-head attention (MHA) framework for yielding a high-quality predictive analytics model. The VIB criteria enables ADEPT to apply a filtering mechanism that relies on generating information-rich text embedding representations that can be used to boost classification accuracy. As a consequence of the VIB criteria, ADEPT can be directly applied on RFRs of timeseries inputs \underline{without any prior preprocessing}. We note that the ADEPT framework is generally extensible and can be used in conjuction with other learning paradigms such as MLPs or SVMs. Figure~\ref{fig:overview} presents a comparative overview of the capabilities of existing time-series classification frameworks versus the proposed \textit{ADEPT} framework. Our contributions can be summarized as follows:
\vspace{-2mm}
\begin{itemize}[noitemsep, topsep=0pt, leftmargin=*, labelsep=0.5em]
    \item We propose \textit{ADEPT}, the first framework to leverage text embeddings for learning representations directly from raw time series data, enabling an automated data engineering pipeline process that is robust to data integrity issues, without finetuning or preprocessing.
    \item ADEPT further enhances these representations by leveraging the VIB criteria, which filters out noise and maximizes application-specific information extracted from the embeddings. 
    \item ADEPT framework integrates of text embedding models along with VIB criteria and transformer based attention models to achieve a fully end-to-end time-series classification pipeline that rivals the performance of state-of-the-art models across diverse applications.
\end{itemize}

\section{Background}
In this section, we formalize multivariate, multi-view classification settings that motivate ADEPT framework design, provide a review of traditional pipeline-based approaches, AutoML strategies, and recent efforts to apply text embeddings in non-text domains before introducing the information-theoretic principles for variational information bottleneck to distill noise and redundancy. 
\subsection{Multivariate Time Series Classification}
We consider a multivariate time series classification problem on a dataset of \(N\) samples or events, where each event \(i\in\{1,\dots,N\}\) has an associated class label \(y_i\in\{1,\dots,C\}\) and can be observed under up to \(K\) temporal views. 
Each temporal view is representative of a specific historical window of timeseries data preceding the occurrence of event $i$. Additionally, each \emph{view} \(k\) denotes a fixed window relative to some anchor point (e.g., \(k=1\) for the earliest window, \(k=K\) for the most recent). Let $S_{i,k} \;\in\; \mathbb{R}^{T\times F}$ be the multivariate time series for view \(k\) corresponding to event \(i\) such that $S_{i,k}$ consists of \(T\) timesteps and \(F\) channels. In other words, we can represent the timeseries data for $T \times K$ consecutive time steps immediately preceding the occurrence of sample $i$ using $S_i$, where $S_i \in \mathbb{R}^{(T\times K)\times F}$ and $S_i = \texttt{Concatenate}(S_{i,k})^{K}_{k=1}$. While this formulation unifies diverse domains, practical pipelines typically transform $S_i$ into fixed-length feature vectors via manual, domain-specific steps (see next subsection). In domains like internet of things (IoT) and predictive maintenance—where several pre-event windows are available—it is natural to use \(K>1\); in other settings, such as EEG analysis, only a single time window is recorded, so one simply sets \(k=1\).

\subsection{Conventional Time Series Classification Pipeline}
Conventional pipelines transform each raw series \(S_{i,k}\) into features via a chain of operations that are finetuned in an iterative cycle of finetuning. This process involves a series of operations, such as:
\begin{equation}
\label{CPA}
\tilde S_{i,k} = f_{\mathrm{imp}}(S_{i,k};\theta_{\mathrm{imp}}),\quad
\hat S_{i,k} = f_{\mathrm{norm}}(\tilde S_{i,k};\theta_{\mathrm{norm}}),\quad
x_i = \phi(\{\hat S_{i,k}\}_{k=1}^K;\theta_{\mathrm{feat}})\in\mathbb{R}^M
\end{equation}
Here, \(f_{\mathrm{imp}}\) denotes missing‐value interpolation (e.g.\ cubic‐spline~\cite{mckinley1998cubic} or Gaussian‐process~\cite{roberts2013gaussian}), \(f_{\mathrm{norm}}\) represents normalization (min–max or \(z\)‐score), and \(\phi\) extracts \(M\) handcrafted features (often \(M\gg TFK\); e.g.\ TSFEL yields \(M>9000\)~\cite{barandas2019tsfel}). Each operation depends on tuning parameters—outlier‐detection thresholds \(\theta_{\mathrm{out}}\)~\cite{hodge2004survey,leys2013detecting}, dimensionality reduction technique like PCA component counts~\cite{keogh2001dimensionality}, or feature‐selection heuristics~\cite{li2017feature}—leading to iterative cycles of hypothesis and validation that can occupy analysts for weeks~\cite{tawakuli2024survey}. Inherently, a successful pre‐processing and feature generation are inherently domain‐specific and demand substantial engineering time and effort to tailor methods and validate results \cite{keogh2001dimensionality,leys2013detecting,tawakuli2024survey,li2017feature,leys2013detecting}. 
Once these preprocessing and data‐engineering steps produce the fixed‐length feature vectors \(\{x_i\}\), practitioners then train a classifier \(g(x_i;\theta_{\mathrm{clf}})\) to predict labels \(y_i\), further extending the design burden with choices of model family and hyperparameters. \textit{Conventional methods require significant engineering time, produce highly complex frameworks, do not generalize well across different applications and problem settings, and often require extensive retraining and parameter re‐tuning when deployed in new environments.}

\subsection{AutoML-Based Pipeline Search for Time Series Classification}
To alleviate the intensive engineering effort of manual preprocessing and feature engineering, AutoML frameworks seek to automate parts of the process by formulating the problem using a joint optimization formulation over a pipeline search space \(\mathcal H\):
\begin{equation}
\label{AutoML}
H^* \;=\; \arg\min_{H\in\mathcal H}
\mathcal L_{\mathrm{val}}\bigl(\mathcal M(\{S_{i,k}\};H)\bigr)
  \end{equation}
where \(\mathcal M\) encompasses imputation, normalization, feature extraction, model architecture, and hyperparameters. Tools such as auto-sklearn’s Bayesian optimization~\cite{feurer2015efficient}, TPOT’s genetic programming~\cite{olson2016evaluation}, AutoKeras’s neural architecture search~\cite{jin2023autokeras}, H2O AutoML’s stacked ensembles~\cite{ledell2020h2o}, and AutoLDT’s CMA-ES–driven transformer search~\cite{wang2024autoldt} have demonstrated the feasibility of AutoML methods.

However, AutoML approaches also incur substantial computational overhead due to the combinatorial size of \(\mathcal H\), often demanding days of GPU/CPU time; they produce opaque “black‐box” pipelines that hinder model interpretability; they typically employ only generic imputation (e.g.\ mean/median) and scaling routines rather than domain-specific methods such as Gaussian‐process interpolation or seasonality‐aware normalization; they still rely on extensive manual filtering of large feature sets (e.g.\ pruning TSFEL’s thousands of extracted features~\cite{barandas2019tsfel}); and their domain-agnostic search strategies frequently overlook temporal inductive biases and multi-view patterns, which can lead to suboptimal accuracy on complex sequence data. \textit{In summary, while AutoML methods improve automation, they do not necessarily lead to a good representation of data, still require significant engineering time, and demands access to immense computational resources.}

\subsection{Text Embedding Models for Domain-Specific Data analysis}
Text embedding models—originally grounded in the distributional hypothesis \cite{harris1954distributional} —treat any co-occurring entities as “tokens,” unlocking cross-domain applications across spatial semantics \cite{hu2020framework,niu2021delineating}, movement dynamics \cite{murray2023unsupervised}, behavioral inference \cite{richie2019predicting}, political discourse analysis \cite{rheault2020word}, joint video–text embeddings for instructional content \cite{sun2019videobert,miech2019howto100m} and audio–text alignments via contrastive pretraining \cite{guzhov2022audioclip,ilharco2019large}. These foundational studies trained embeddings from scratch on domain-specific data, demonstrating that lightweight embedding architectures can effectively capture complex, domain-specific structures. Building on this legacy, modern practitioners can either deploy fully offline, open-source embedders—such as Nomic’s nomic-embed-text-v1 \cite{nussbaum2024nomic}—for strict data residency and privacy control, or leverage API-based services like OpenAI’s text-embedding-3-small \cite{openai_text_embedding_3_small_2023}, which often offer superior accuracy due to web-scale pretraining but require sending inputs to third-party servers. Because these large models are pretrained on vast, internet-scale corpora, they provide high-quality semantic vectors with far less overhead than full LLMs, streamlining experimentation without building custom models from scratch. \textit{Text embedding models demonstrated a significant ability to capture structure across diverse domains —suggesting that even highly structured, non-linguistic data like time series may benefit from such pretrained semantic representations.}

\subsection{Information Bottleneck Approaches in Deep Learning}

Variational Information Bottleneck (VIB) techniques emerged as powerful tools in deep learning to denoise input data and enhance model accuracy. The original formulation by \cite{alemi2016deep} introduced a stochastic encoder–decoder framework that improved model robustness on MNIST and ImageNet by compressing task-irrelevant features in the latent space. This was extended by \cite{achille2018information}, who proposed Information Dropout—a parameterized log-normal noise model that promotes invariant, disentangled representations. 
Foundational insights from noisy channel theory \cite{dobrushin1962information} support VIB’s core mechanism of stochastic compression. In generative modeling, the $\beta$-VAE of \cite{higgins2017beta} similarly enforces factorized, expressive latent codes via a constrained variational objective. A comprehensive survey by \cite{goldfeld2020information} synthesizes these developments, framing VIB as a unifying principle across representation learning. \textit{By suppressing redundancy and irrelevant noise, VIB generates more informative and compact latent representations, and ultimately improves the performance of the downstream prediction tasks.}




\section{Methodology}
The methodological core of ADEPT relies purely on a black box language model applied on decompositions of the RFRs of timeseries input datasets which is followed by a VIB criterion for enhancing information gain. The ADEPT methodology can be broken down into four distinct steps that can be implemented in a scalable fashion, and integrated to develop two versions of the framework.
\vspace{-1mm}
\begin{itemize}[noitemsep, topsep=0pt, leftmargin=*, labelsep=0.5em]
\item \emph{RFR Processing and Decomposition}: RFRs corresponding to multichannel timeseries input sequences are decomposed into segments of fixed content sizes and serialized for standardization.
\item \emph{Temporal Text Embeddings}: Serialized RFR temporal decompositions are processed using black-box, LLM-based language embedding models to obtain text embeddings.
\item \emph{Variational Information Bottleneck}: A variational encoder learns the latent space distribution of RFR embeddings, resulting in fused sequences to reduce noise \& enhance information gain.
\item \emph{Classifier}: A transformer-based classifier captures intra- and inter-view dependencies from the fused multi-view sequences and performs final prediction.
\end{itemize}


\subsection{RFR Processing and Decomposition}\label{sec:phase1}
We begin by considering the tuple $(R_i,y_i)$ for each reported event $i\in \{1,\ldots,N\}$, where $R_i = \texttt{RFR}(S_i)$ where $S_i \in \mathbb{R}^{(T.K)\times F}$ denotes the actual timeseries data for $K$ temporal views immediately preceding event $i$. Next, we consider the decomposition of each temporal view $S_{i,k}$ into \(M\) equal‐length segments or chunks of length $L=T/M$, with corresponding RFR $R^{(j)}_{i,k} = \texttt{RFR}(S^{j}_{i,k})$. It is important to note that the decomposition scheme preserves the temporal order of data pertaining to individual time steps across as well as within multiple views. Therefore, extracting the RFR for each segment can be trivially accomplished using a simple count based query or by enforcing a content size limit (for eg: in KBs, MBs) on each chunk. 

Our approach also balances the extremes of processing the full \(T\times F\) series at once—which can dilute important local patterns and incur high computational cost—and treating each timestep independently—which ignores temporal and cross‐channel structure. While temporal chunking enhances downstream representations by balancing local and global dependencies, selecting the optimal number of chunks \(M\) introduces a trade‐off.  A smaller \(M\) (longer chunks) may overload downstream encoders or mix heterogeneous patterns, whereas a larger \(M\) (shorter chunks) risks fragmenting temporal dependencies and increasing sequence length. Domain insight or systematic validation studies can help inform the choice of \(M\) for balancing expressivity and computational tractability. However, validation studies for determining \(M\) can be easily automated and implemented in a scalable fashion on account of the pipeline simplifications afforded by the text embedding models.

\subsection{Temporal Text Embeddings}
\label{sec:phase4}
To leverage powerful, pre‐trained semantic priors, we treat each raw time‐series chunk as text, enabling off‐the‐shelf embedding models to capture both numeric and categorical patterns without manual intervention. We serialize each RFR chunk \(R_{i,k}^{(j)}\in\mathbb{R}^{L\times F}\) to obtain \(R_{i,k}^{(j),ser} \;\in\; \Sigma^*\) by concatenating timestamps and channel readings into a token sequence where \(\Sigma^*\) is the model’s character set and \(E\) its output dimension. We then apply a frozen text‐embedding function \(g\):
\vspace{-1mm}
\begin{equation}
\label{pure embedding3}
\mathbf{e}_{i,k}^{(j)} = g\bigl(R_{i,k}^{(j),ser}\bigr) \;\in\; \mathbb{R}^{E}
\end{equation}
\textbf{Note:} Since \(g\) natively handles both numeric and textual tokens, categorical channels (e.g., flags or event types) can be embedded alongside continuous measurements in one unified string.
 
Pre‐trained text embeddings—trained on general LLM corpora—can introduce noise when representing precise numerical sequences and often yield very high‐dimensional, redundant vectors. To address these issues, we apply a Variational Information Bottleneck in the next step to distill more informative, lower‐dimensional representations.  

\subsection{Variational Information Bottleneck Criteria}
\label{sec:phase2}
To reduce noise variance, redundancy and maximize the extracted information gain from high‐dimensional text embeddings \(\mathbf{e}_{i,k}^{(j)}\), we adopt a VIB criterion across each view, producing compact low-dimensional encodings that retain task‐relevant information. To do so, we compress \(\mathbf{e}_{i,k}^{(j)}\) into a \(d\)-dimensional code \(\mathbf{z}_{i,k}^{(j)}\) by leveraging a VIB encoder \cite{alemi2016deep}. For each view \(k\), let \(\phi_k=\{W_{\mu}^{(k)},b_{\mu}^{(k)},W_{\log\varphi}^{(k)},b_{\log\varphi}^{(k)}\}\) denote the VIB encoder parameters, and \(\theta_k=\{W_{y}^{(k)},b_{y}^{(k)}\}\) the linear classifier parameters.  Here \(d\) is the \emph{bottleneck dimension} and \(\beta>0\) the VIB trade-off weight. 

\noindent\textbf{Stochastic encoder for view \(k\)}: For each chunk chunk \(\mathbf{e}_{i,k}^{(j)}\), we derive a low dimensional representation defined by $\mu_k$, $\sigma_k$ as defined in Equation \eqref{eq:stochen}.
 \begin{equation}\label{eq:stochen}
 \vspace{-1mm}
   \mu_k = W_{\mu}^{(k)}\,\mathbf{e}_{i,k}^{(j)} + b_{\mu}^{(k)}, 
\quad
\log\sigma_k^2 = W_{\log\varphi}^{(k)}\,\mathbf{e}_{i,k}^{(j)} + b_{\log\varphi}^{(k)}
  \end{equation}
In Equation \eqref{eq:stochen}, we clip $\sigma_k$ to lie between \([-10,10]\), and set \(\sigma_k=\exp\bigl(\tfrac12\log\sigma_k^2\bigr)\). As a result, the encoder can be used to represent the conditional latent space distribution based on the observed embeddings as given in Equation \eqref{stochastic2}.
 \begin{equation}
  \label{stochastic2}
   q_{\phi_k}(z\mid \mathbf{e}_{i,k}^{(j)})
= \mathcal{N}\bigl(z;\mu_k,\mathrm{diag}(\sigma_k^2)\bigr)
  \end{equation}

\noindent\textbf{Reparameterization trick}: In order to learn the latent space distribution conditioned on the embeddings, we apply a reparametrization trick characterized by Equation \eqref{reparameter} where \(\varepsilon\sim\mathcal{N}(0,I_d)\).
\begin{equation}
  \label{reparameter}
   \mathbf{z}_{i,k}^{(j)} = \mu_k + \sigma_k \,\odot\, \varepsilon
  \end{equation}
The reparametrization trick ensures that the distribution of latent space can be parametrized by \(\phi_k\) which can be learned using gradient descent.

\noindent\textbf{Classification head}: We augment the VIB stochastic encoder with a classification head that is parametrized by $\theta_k$ in order to map \(\mathbf{z}_{i,k}^{(j)}\) to logits represented in Equation \eqref{classification}. 
\begin{equation}
  \label{classification}
\ell_{i,k}^{(j)} = W_{y}^{(k)}\,\mathbf{z}_{i,k}^{(j)} + b_{y}^{(k)},
\quad
p_{\theta_k}(y_i\mid \mathbf{z}_{i,k}^{(j)}) = \mathrm{Softmax}(\ell_{i,k}^{(j)})
  \end{equation}
\noindent\textbf{Per-view loss}: In order to jointly train the stochastic encoder and decoder framework, we adopt a loss function characterized by Equation \eqref{loss}.
\begin{equation}
  \label{loss}
\mathcal{L}_k
= \frac{1}{NM}\sum_{i=1}^N\sum_{j=1}^M
\Bigl[-\log p_{\theta_k}(y_i\mid \mathbf{z}_{i,k}^{(j)})\Bigr]
\;+\;\beta\,\frac{1}{NM}\sum_{i,j}
D_{\mathrm{KL}}\!\bigl[q_{\phi_k}(z\mid \mathbf{e}_{i,k}^{(j)})\;\|\;\mathcal{N}(0,I_d)\bigr]
  \end{equation}
where $D_{\mathrm{KL}}\!\bigl(\mathcal{N}(\mu,\sigma^2)\|\mathcal{N}(0,I)\bigr)
= \tfrac12\sum_{\ell=1}^d\bigl(\mu_\ell^2+\sigma_\ell^2-\log\sigma_\ell^2-1\bigr)$ represents the Kullback-Leibler divergence loss. 
The cross-entropy term ensures each view’s encoder retains predictive information; the KL term, 
enforces compact, robust codes. We optimize each \(\mathcal{L}_k\) independently via Adam (learning rate \(\eta\), batch size \(B\)) for \(E\) epochs, yielding specialized encoder–classifier parameters \((\phi_k,\theta_k)\) for $k$.  
\subsection{Transformer-based Classifier Design}
\label{sec:phase6}
For the final prediction task, we train a classifier on the latent space variates $\mathbf{z}_{i,k}^{(j)}$.  While any classifier (e.g.\ SVM, random forest, XGBoost) could be used, we adopt a powerful multi‐head attention (MHA) based Transformer to capture both intra‐ and inter‐view dependencies in a unified model. In the step, we train a single Transformer‐based model on the fused embeddings from all $k$ views to perform final classification.  For each sample \(i\), view \(k\), and chunk \(j\in\{1,\dots,M\}\), let $\mathbf{z}_{i,k}^{(j)} \in \mathbb{R}^{E}$ be the embedding.  We assemble these into a sequence $F_{i,k} = \bigl[\mathbf{z}_{i,k}^{(1)}, \mathbf{z}_{i,k}^{(2)}, \dots, \mathbf{z}_{i,k}^{(M)}\bigr]^\mathsf{T}
\;\in\;\mathbb{R}^{M\times E}$. Our goal is to predict the class label \(y_i\in\{1,\dots,C\}\) for each sample \(i\), leveraging all k views. Therefore, we 
define a TransformerAutoencoder \(T_\psi\) architecture with parameters \(\psi\) on the following components.

\noindent\textbf{Input projection and positional encoding}: A linear layer projects each \(E\)-dimensional row of \(F_{i,k}\) into a \(h\)-dimensional latent space (Equation \eqref{input projection}).
\begin{equation}
\label{input projection}
H^{(0)}_{i,k} = F_{i,k}\,W_{\mathrm{in}} + b_{\mathrm{in}},
\quad W_{\mathrm{in}}\in\mathbb{R}^{E\times h},\,b_{\mathrm{in}}\in\mathbb{R}^h
\end{equation}
Consequently, the learned embeddings are added to each timestep so as to inject order information.
 
\noindent\textbf{Transformer Encoder \& Decoder}: Stacked multi‐head self‐attention layers encode the sequence into \(H^{(L)}_{i,k}\in\mathbb{R}^{M\times h}\), then a decoder reconstructs \(\hat F_{i,k}\in\mathbb{R}^{M\times E}\).

\noindent\textbf{Classification head}: We pool \(H^{(L)}_{i,k}\) over time,
\(\bar h_{i,k} = \tfrac1M\sum_{j=1}^M H^{(L)}_{i,k,j}\in\mathbb{R}^h,\)
and map to class logits as given in Equation \eqref{lhead} which can be used to compute probabilities \(p_{i,k}=\mathrm{Softmax}(\ell_{i,k})\in\Delta^{C}\).
\begin{equation}
  \label{lhead}
  \ell_{i,k} = \bar h_{i,k}\,W_{y} + b_{y},
\quad W_{y}\in\mathbb{R}^{h\times C},\,b_{y}\in\mathbb{R}^C
\end{equation}


\noindent\textbf{Training procedure}: We first \emph{pretrain} \(T_\psi\) as a joint autoencoder by minimizing the mean‐squared reconstruction error averaged across the k views as represented in Equation \eqref{training}.
\begin{equation}
    \label{training}
    \mathcal{L}_{\mathrm{AE}}
= \frac{1}{K.N}\sum_{i=1}^N\sum_{k=1}^K \frac{1}{2M.E}\bigl\|\hat F_{i,k} - F_{i,k}\bigr\|_F^2
\end{equation}
This encourages the model to learn a latent representation that reconstructs all fused embeddings, capturing common structure across views. Next, we \emph{fine‐tune} for classification by computing the per‐view distributions \(p_{i,1},p_{i,2},p_{i,3}\) for each sample \(i\), leading to a consensus given by Equation \eqref{fine tune}.
\begin{equation}
    \label{fine tune}
    p_i = \frac{p_{i,1}\,\odot\,p_{i,2}\,\odot\,p_{i,3}}
{\sum_{c=1}^C \bigl[p_{i,1}\,\odot\,p_{i,2}\,\odot\,p_{i,3}\bigr]_c}
\end{equation}
and minimize the negative log-likelihood loss $\mathcal{L}_{\mathrm{NLL}}
\;=\;
-\,\frac{1}{N}\,\sum_{i=1}^{N}\log p_{i}[y_{i}]$.


\subsection{ The ADEPT Framework}
\begin{figure}[!htbp]
  \centering  \includegraphics[width=1.0\textwidth]{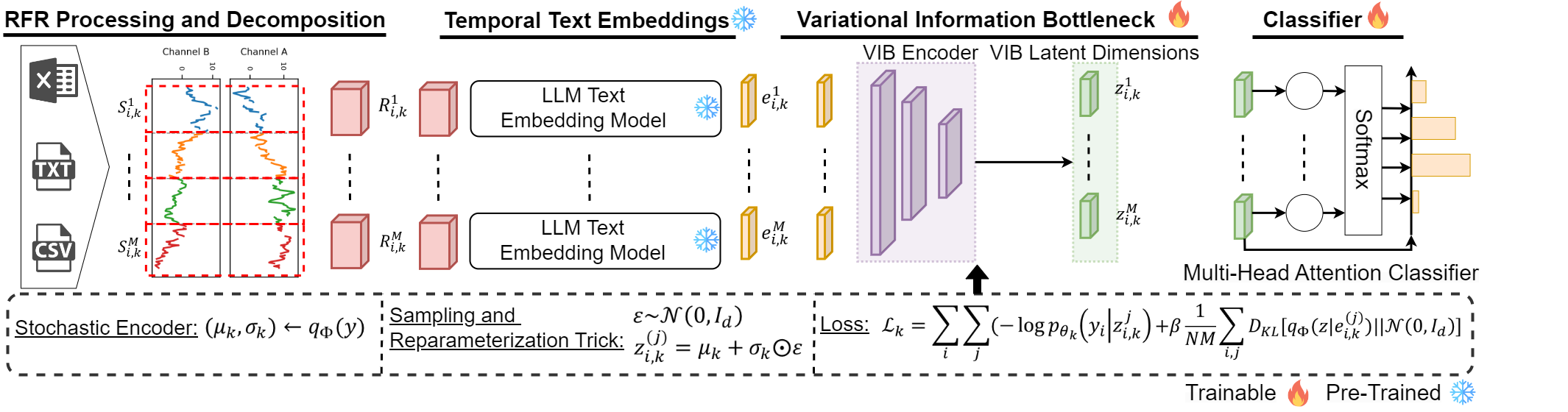}
  \caption{Illustration of the ADEPT v2.0. Framework
  }
  \label{fig:how}
\end{figure}

We define two versions of the \textit{ADEPT} framework. ADEPT v1.0 is the baseline version of our pipeline, 
which directly applies pretrained text embeddings to serialized time-series segments, followed by a multi-head attention classifier. ADEPT v2.0 extends this baseline by incorporating a VIB layer, which compresses the raw embeddings into compact, task-relevant codes that reduce noise and improve generalization. While both versions eliminate the need for traditional data engineering steps, ADEPT v2.0 introduces an additional mechanism to better align learned representations with downstream prediction objectives. The general pipeline of ADEPT v2.0 is shown in Figure~\ref{fig:how} and Algorithm~\ref{alg:adept_v2_text}.  
\begin{algorithm}[H]
\footnotesize
\caption{ADEPT v2.0}
\label{alg:adept_v2_text}
\begin{algorithmic}[1]
\Require Series $S_{i,k}$, embedding model $g$, VIB params $\phi_k,\theta_k$, Transformer‐AE params $\psi$, hyperparams $(\beta,\eta,\eta^{\prime},\eta^{\prime \prime},E_{\mathrm{VIB}},E_{\mathrm{AE}},E_{\mathrm{CL}})$  
\State Partition $S_{i,k}$ into $M$ segments of length $L=\frac{T}{M}$, $\forall i \in [N], \forall k \in [K]$. \Comment{RFR Decomposition}
\State Compute embeddings $\mathbf{e}_{i,k}^{(j)}$ (Eq.~\ref{pure embedding3}) using text embedding model $g$. \Comment{Temporal Text Embeddings}  
\For{$k=1,\dots,K$} \Comment{Variational Information Bottleneck}
  \For{epoch $=1,\dots,E_{\mathrm{VIB}}$}
    \State Compute the per‐view VIB loss $\mathcal{L}_k$ using (Eqs.~\ref{loss}) and update $\phi_k,\theta_k$ via Adam($\eta$).
  \EndFor
  \State Utilize updated $\phi_k,\theta_k$ to obtain latent space encodings $\{\mathbf{z}_{i,k}^{(j)}\}$. 
\EndFor
\For{epoch $=1,\dots,E_{\mathrm{AE}}$} \Comment{Transformer‐AE pretraining}
  \State Assemble each view’s latent encodings, compute loss $\mathcal{L}_{AE}$ (Eq.~\ref{training}) and update $\psi$ via Adam($\eta^{\prime}$).
\EndFor
\For{epoch $=1,\dots,E_{\mathrm{CL}}$}\Comment{Classification fine‐tuning}
    
  \State Compute per‐view class probabilities $p_{i,k} = \mathrm{Softmax}\bigl(T_\psi(F_{i,k})\bigr)$, $\forall k \in K$.
  \State Fuse the per‐view probabilities into a final distribution $p_i$ (Eq.~\ref{fine tune}).
  \State Compute negative log‐likelihood loss $\mathcal{L}_{NLL}=\frac{-1}{N}\,\sum\limits_{i=1}^{N}\log p_{i}[y_{i}]$ and update $\psi$ via Adam($\eta^{\prime \prime}$).
\EndFor
\State Use Transformer‐AE $\psi$ to obtain fused probabilities $p_i$ for predicting $\displaystyle \hat y_i = \arg\max_c\,p_i[c]$ \Comment{Inference}
\end{algorithmic}
\end{algorithm}

\vspace{-1mm}
\section{Experimental Results}
This section evaluates the predictive performance of the proposed ADEPT framework across datasets drawn from diverse domains, including healthcare, science, finance, and IoT. These datasets introduce a large variety of challenges such as data integrity, temporal dependencies, and privacy constraints. The associated prediction tasks also vary significantly in complexity—from ternary classification of Bitcoin price direction to multi-class root cause analysis in hydroelectric systems, where each failure mode is represented by only a handful of observations. To tackle these domain-specific challenges, existing approaches rely on heavily customized data engineering pipelines, manually optimized for each dataset and task. 
In our experiments, we benchmark ADEPT against these application-specific state-of-the-art models. 
Details of the datasets, evaluation metrics, and implementation are provided in Appendices~\ref{dataset},~\ref{evaluation metrics}, and~\ref{implementation}, respectively.
\subsection{Experiments on a Science Application - \textit{Predicting the Astrophysical Class of Light Curves}}

The science application experiment is from an astrophysics example ( PLAsTiCC dataset from the \textit{2018} Kaggle competition) that aims to predict the astrophysical class of light curves. The dataset consists of $7,848$ simulated time series LSST light curves labeled across $14$ astrophysical classes. The data, with variable-length sequences and $~30\%$ missing per-band flux values, is split into training, validation, and test sets. Table~\ref{tab:lsst_classification} compares our two frameworks—\emph{ADEPT v1.0} (without VIB) and \emph{ADEPT v2.0} (with VIB)—against three state-of-the-art classifiers that yield the best accuracies in literature: CATS~\cite{fraga2024transient}, AMPEL~\cite{nordin2025ampel}, and ORACLE~\cite{shah2025oracle}. The bechmark models employ conventional time-series pipelines with multiple preprocessing steps such as filtering, normalization, truncation, and padding; and leverage specialized architectures like CNN+LSTM hybrids or hierarchical RNNs to extract temporal features from the multi-band photometric data. Their reported classification accuracies range from 80\% to 84\%, which signifies the difficulty of the prediction task. Unlike these methods — which rely on hand-crafted time and color features followed by gradient-boosted trees or hierarchical RNNs — our pipelines operate directly on text-serialized light curves. \emph{ADEPT v1.0} achieves 95.98\,\% accuracy, while incorporation of VIB in \emph{ADEPT v2.0} further improves the accuracy to 97.83\,\%, outperforming all benchmarks with an improvement of >10\%.
\captionsetup[subtable]{skip=2pt} 
\begingroup
\small 
\begin{table}[!htbp]
  \centering
  \caption{Benchmarking Results Across a Range of Applications}
  \label{tab:benchmarking_combined}
  \begin{subtable}[t]{\textwidth}
    \centering
    \caption{Science Dataset: \textit{PLAsTiCC Classification}}
    \label{tab:lsst_classification}
    \begin{tabular}{@{}p{2.1cm}p{7.2cm}p{2.5cm}c@{}}
      \toprule
      \textbf{Source} & \textbf{Preprocessing Steps} & \textbf{Classifier} & \textbf{Accuracy} \\
      \midrule
      CATS\cite{fraga2024transient} & \textit{Clean data, filter,  normalize, etc.} & CNN+LSTM & 83\% \\
      AMPEL\cite{nordin2025ampel} & \textit{Filter, remove noise, negative-flux, sampling, etc.} & ParSNIP + GBM & 80\% \\
      ORACLE\cite{shah2025oracle} & \textit{Remove noise, truncate, pad, normalize, mask, etc.} & Hierarchical RNN 
      & 84\% \\
      \emph{ADEPT v1.0} & \textit{Bypassed via text embedding} & MHA & 95.98\% \\
      \textbf{\emph{ADEPT v2.0}} & \textbf{\textit{Bypassed via text embedding $+$ VIB}} & \textbf{MHA} & \textbf{97.83\%} \\
      \bottomrule
    \end{tabular}
  \end{subtable}
  \begin{subtable}[t]{\textwidth}
    \centering
    \caption{Healthcare Dataset: \textit{SelfRegulationSCP2 Classification}}
    \label{tab:SelfRegulationSCP2_classification}
    \begin{tabular}{@{}p{2.1cm}p{7.2cm}p{2.5cm}c@{}}
      \toprule
      \textbf{Source} & \textbf{Preprocessing Steps} & \textbf{Classifier} & \textbf{Accuracy} \\
      \midrule
      MiniRocket\cite{keshavarzian2023representation} & \textit{Augment via random freq. butchering, etc.} & MiniRocket & 59.0\% \\
      MHCAN\cite{huang2024multilevel} & \textit{Multilevel DWT, 1D-Conv, positional encoding, etc.} & MHCA & 62.20\% \\
      TSEM\cite{pham2023tsem} & \textit{2D-Conv filters, LSTM, spatiotemporal maps, etc.} & Transformer & 75.60\% \\
      \emph{ADEPT v1.0} & \textit{Bypassed via text embedding} & MHA & 58.97\% \\
      \textbf{\emph{ADEPT v2.0}} & \textbf{\textit{Bypassed via text embedding $+$ VIB}} & \textbf{MHA} & \textbf{73.68\%} \\
      \bottomrule
    \end{tabular}
  \end{subtable}
  \begin{subtable}[t]{\textwidth}
    \centering
    \caption{Financial Dataset: \textit{Bitcoin Price Trend Classification}}
    \label{tab:crypto_ts_classification}
    \begin{tabular}{@{}p{2.1cm}p{7.2cm}p{2.5cm}c@{}}
      \toprule
      \textbf{Source} & \textbf{Preprocessing Steps} & \textbf{Classifier} & \textbf{Accuracy} \\
      \midrule
      RLSTM\cite{kwon2019time} & \textit{Forward‐fill missing data, drop outliers, etc.} & LSTM & 66\% \\
      EDL\cite{rao2023time} & \textit{Remove anomaly, normalize, sequence structuring} & CNN–LSTM & 64\% \\      DQN\cite{muminov2024enhanced}  & \textit{Reduce noise, normalize, feature fusion, etc.} & Deep Q-Network & 95\%$^*$ \\
      \emph{ADEPT v1.0} & \textit{Bypassed via text embedding} & MHA & 45.40\% \\
      \textbf{\emph{ADEPT v2.0}} & \textbf{\textit{Bypassed via text embedding $+$ VIB}} & \textbf{MHA} & \textbf{88.49\%} \\
      \bottomrule
    \end{tabular}
    \textit{{$^*$ \underline{\textit{Note}}: The model in \cite{muminov2024enhanced} uses extra features from on-chain metrics, X and Google Trends, which are not available to ADEPT for training.}}
  \end{subtable}
  \begin{subtable}[t]{\textwidth}
    \centering
    \caption{Internet-of-Things Dataset: \textit{Hydropower-Research Institute Fault Classification}}
    \label{tab:acc_hri}
    \begin{tabular}{@{}p{2.1cm}p{7.2cm}c@{}c@{}c@{}}
      \toprule
      \textbf{Model} & \textbf{Preprocessing Steps} & \hspace{0.2cm}\textbf{Accuracy \hspace{0.2cm}} & \textbf{Top-2 Accuracy} \\
      \midrule
      \emph{TSFEL+MHAN} & \textit{TSFEL features, MI selection, normalize, etc.} & 42.80\% & \textbf{57.14}\% \\
      \emph{ADEPT v1.0} & \textit{Bypassed via text embedding} & 45.00\% & \textbf{66.67}\% \\
      \textbf{\emph{ADEPT v2.0}} & \textbf{\textit{Bypassed via text embedding $+$ VIB}} & \textbf{74.35\%} & \textbf{\textbf{97.5}\%} \\
      \bottomrule
    \end{tabular}
  \end{subtable}
\end{table}
\endgroup
\subsection{Experiments on a Healthcare Application - \textit{Predicting Patient Condition
using EEG Data}}

The healthcare application focuses on classification using the SelfRegulationSCP2 dataset, a multivariate time-series dataset derived from electroencephalography (EEG) recordings. The dataset comprises trials labeled across two cognitive classes (positive and negative) related to self-regulated cortical potentials, with each instance representing 7-channel EEG signals recorded over 4.5 seconds. Results are shown in Table~\ref{tab:SelfRegulationSCP2_classification}. For this dataset, we use three high-performing benchmarks from recent literature: MiniRocket~\cite{keshavarzian2023representation}, MHCAN~\cite{huang2024multilevel}, and TSEM~\cite{pham2023tsem}. These methods follow conventional multivariate time-series classification pipelines, using application-specific preprocessing steps such as wavelet decomposition, temporal convolutions, and spatiotemporal mapping, coupled with specialized architectures like transformers and hybrid CNN-RNN models. Their reported classification accuracies range from 59.0\% to 75.60\%, which constitutes a significant spread, showcasing that the capability of inherent indicators of mental state are challenging to discover. 

Among the proposed models, \emph{ADEPT v1.0}, despite its simplicity, achieves a comparable 58.97\% accuracy. With the addition of a variational information bottleneck in \emph{ADEPT v2.0}, accuracy improves to 73.68\%, outperforming two of the three benchmarks and closely approaching the best-performing method. An interesting observation in this experiment is the significant accuracy gap between \emph{ADEPT v1.0} and \textit{v2.0} models. The VIB step in \emph{ADEPT v2.0} unlocks a richer representation of data and results in 14.7\% improvement in accuracy.


\subsection{Experiments on a Finance Application - \textit{Predicting Future Bitcoin Price Trend}}

The financial application focuses on next-day trend classification using the Bitcoin Price Trend Dataset, which spans daily BTC/USD OHLCV records and 14 technical indicators over the period 2015–2023. The task is framed as a 3-class classification problem—predicting whether the price will rise by more than 1\%, fall by more than 1\%, or remain stable within 1\%. We train and tune models on data through 2022 and evaluate performance on all 365 trading days of 2023. Table~\ref{tab:crypto_ts_classification} compares our framework, \emph{ADEPT v1.0} and \emph{ADEPT v2.0}, against three baseline methods from the literature: a Recurrent LSTM~\cite{kwon2019time}, a Ensemble Deep Learning~\cite{rao2023time}, and a Deep Q-Network~\cite{muminov2024enhanced}. These baselines rely on traditional financial preprocessing steps including normalization, anomaly removal, and hand-engineered feature selection, with accuracies ranging from 64\% to 95\%. 

In this application, \emph{ADEPT v1.0} achieves a poor accuracy of 45.40\%. However, incorporating VIB in \emph{ADEPT v2.0} significantly boosts performance to 88.49\%. This is the best performing model across benchmarks that have access to the same data. The only model that outperforms \emph{ADEPT v2.0} is the model in \cite{muminov2024enhanced}, which has access to additional data from on‐chain metrics, X and Google Trends.

\subsection{Experiments on an IoT Application - \textit{Predicting the Cause of Hydropower Reliability Issues}}
IoT application focuses on predicting the root cause of reliability issues in hydropower components using a proprietary commercial dataset shared with the authors through the courtesy of the Hydropower Research Institute. The original dataset encompasses information from 197 hydropower plants and 844 generating units, which accounts for approximately 42\% of U.S. capacity. The data includes operational metrics, and event logs. We construct the training dataset by aggregating a set of reliability events into a database that pairs each event’s cause code with multi-stream sensor readings captured over several days leading up to the event. There are $14$ unique cause codes. The objective is to predict the right cause code subject to the inherent variability of industrial data due to highly dynamic and heterogeneous conditions, which introduces substantial complexity. 

Since this is proprietary data, there is no prior work on prediction in this dataset. We take this opportunity to test the performance of the existing AutoML methods, specifically the TSFEL package \cite{barandas2020tsfel}. An additional challenge comes from the privacy and data residency requirements for this dataset preventing the use of public LLM-based embedding models such as OpenAI's \texttt{text-embedding-3-small}. Therefore, we leverage the \texttt{nomic-embed-text-v1} model (768\text{-}dim) hosted on-prem to embed each text segment. Details of the implementation for this dataset is provided in Appendix~\ref{implementation}. Table~\ref{tab:acc_hri} reports overall test accuracy: Feature extraction + MHAN achieves 42.80\%, ADEPT v1.0 45.00\%, and ADEPT v2.0 attains 74.35\%. This substantial gain showcases the advantage of combining semantic embeddings with VIB filtering over hand‐crafted features or unfiltered embeddings.


In many IoT-enabled asset monitoring applications, analyzing the performance of the model’s second-best prediction—referred to as Top-2 Accuracy—can be particularly valuable. Top-2 predictions offer actionable insights by identifying plausible alternative failure modes, which can guide proactive inspections. 
This is particularly important in high-stakes settings, such as hydropower systems, where Top-2 alerts can trigger early intervention to prevent failures of heavy assets like turbines or thrust bearings, potentially saving millions of dollars per incident. On this metric, \textit{ADEPT v2.0} has an accuracy of $97.5\%$, which offers significant improvements compared to the benchmarks and the existing commercial solutions.

\subsection{A note on the comparison of ADEPT v1.0 and v2.0}
We note that the ADEPT v2.0 outperforms v1.0 across all the experiments. This is a direct consequence of the VIB step in ADEPT v2.0 removing non-informative and noisy components from the raw text embeddings through a KL-penalized compression during training, which encourages each code $z^{(j)}_{i,k}$ to adhere to a standard normal prior, producing compact codes that align more closely with class-specific patterns. As a result, ADEPT v2.0 exhibits tighter within-class clustering and clearer between class separation-translating into higher accuracy and stronger generalization-whereas v1.0 must contend with the full high-dimensional noise and redundancy of its embedding.


\begin{figure}[htbp]
  \centering
\vspace{-3mm}  \includegraphics[width=1.0\textwidth]{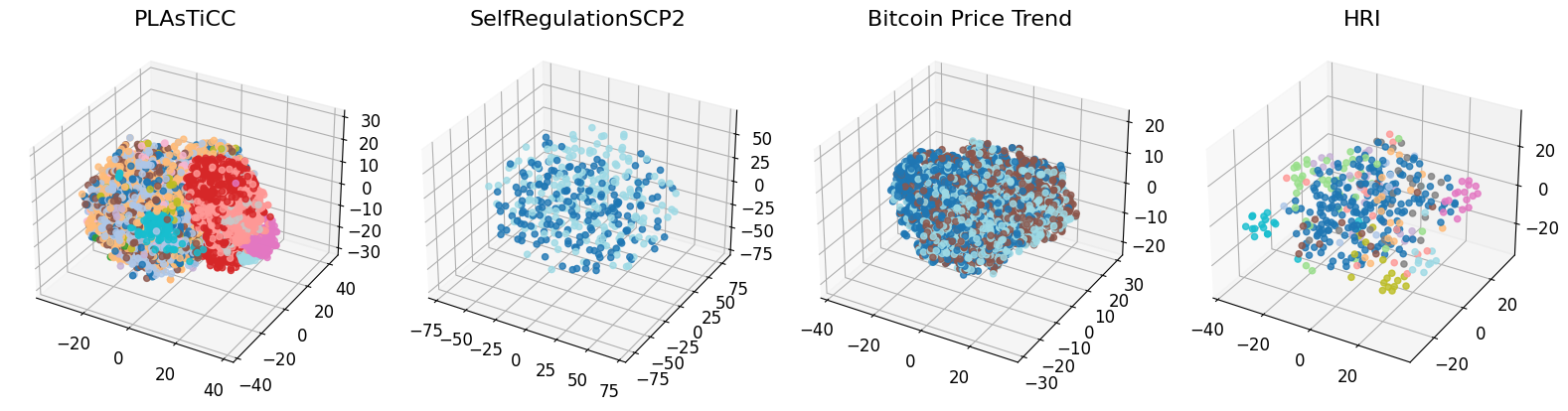}
  \caption{t-SNE projection of segment embeddings (ADEPT v2.0) across different applications.
  }
  \label{fig:all_tsne}
  \vspace{-3mm}
\end{figure}

\section{Conclusion}
We have shown that general-purpose text embeddings—without any additional feature engineering or domain-specific data preprocessing—can serve as powerful representations for raw time‐series classification. Across four diverse experiments (Science, Healthcare, Finance and IoT), ADEPT consistently outperforms application specific engineered predictive models. We demonstrate that off-the-shelf text embedding models, when paired with a lightweight variational information bottleneck step, can capture the salient structure of heterogeneous time-series inputs. This paves the way for fast, turnkey classification solutions in domains where feature engineering is costly or impractical. Experimental results across diverse datasets demonstrate that the proposed \textit{ADEPT v2.0} model consistently matches or surpasses the best-performing benchmarks in all application domains; showcasing that \textit{text embeddings can indeed be highly cost-effective and capable data engineers}.

 
\bibliographystyle{plain}
\bibliography{ref}	

\newpage
\newpage

\appendix
\section{ADEPT v2.0 Algorithm}
\label{alg}
The full pseducode of ADEPT v2.0 is discussed in Algorithm~\ref{alg:full_with_formulas}

\vspace{-4mm}
\begin{algorithm}[H]
\footnotesize
\caption{ADEPT v2.0}
\label{alg:full_with_formulas}
\begin{algorithmic}[1]
\Require 
  Multivariate series $S_{i,k}\in\mathbb{R}^{T\times F}$, chunks $M$,  
  frozen text embedder $g$,  
  VIB params $\phi_k,\theta_k$,  
  Transformer‐AE $T_\psi$,  
  hyperparams $(\beta,\eta,B,E_{\mathrm{VIB}},E_{\mathrm{AE}},E_{\mathrm{CL}})$  
\Ensure 
  Predicted label $\hat y_i$
\medskip

\State \textbf{RFR Processing and Decomposition}
\State $L \gets T/M$
\For{$k=1,\dots,K$}\For{$j=1,\dots,M$}
  \State $S_{i,k}^{(j)} \gets S_{i,k}[(j-1)L+1:jL,\,1:F]$
\EndFor\EndFor

\State \textbf{Temporal Text Embeddings}
\For{$k=1,\dots,K$}\For{$j=1,\dots,M$}
  \State $R_{i,k}^{(j)} = \text{serialize}(S_{i,k}^{(j)})$
  \State $\mathbf{e}_{i,k}^{(j)} = g\bigl(R_{i,k}^{(j)}\bigr)\in\mathbb{R}^E$
    \Comment{(Eq.~\ref{pure embedding3})}
\EndFor\EndFor

\State \textbf{Variational Information Bottleneck}
\For{$k=1,\dots,K$}\For{$j=1,\dots,M$}
  \State $\mu_k = W_{\mu}^{(k)}\mathbf{e}_{i,k}^{(j)} + b_{\mu}^{(k)},$
  \quad $\log\sigma_k^2 = W_{\log\varphi}^{(k)}\mathbf{e}_{i,k}^{(j)} + b_{\log\varphi}^{(k)}$
    \Comment{(Eq.~\ref{eq:stochen})}
  \State $\sigma_k = \exp\bigl(\tfrac12\log\sigma_k^2\bigr)$
  \State Sample $\varepsilon\sim\mathcal{N}(0,I_d)$
  \State $\mathbf{z}_{i,k}^{(j)} = \mu_k + \sigma_k \odot \varepsilon$
    \Comment{(Eqs.~\ref{stochastic2}, \ref{reparameter})}
  \State $\ell_{i,k}^{(j)} = W_{y}^{(k)}\mathbf{z}_{i,k}^{(j)} + b_{y}^{(k)},$
  \quad $p_{\theta_k}(y_i\!\mid\!\mathbf{z})=\mathrm{Softmax}(\ell_{i,k}^{(j)})$
    \Comment{(Eq.~\ref{classification})}
\EndFor
  \State Compute per‐view loss
  \[
    \mathcal{L}_k 
    = \frac1{NM}\sum_{i,j}\bigl[-\log p_{\theta_k}(y_i\mid z)\bigr]
      + \beta\,\frac1{NM}\sum_{i,j}
        D_{KL}\!\bigl[q_{\phi_k}(\cdot)\|\mathcal{N}(0,I)\bigr]
  \]
    \Comment{(Eq.~\ref{loss})}
  \State Update $\phi_k,\theta_k$ via Adam($\eta$) for $E_{\mathrm{VIB}}$ epochs
\EndFor

\State \textbf{Transformer-based Classifier Design}
\State Assemble each view’s sequence
$F_{i,k} = [\,\mathbf{z}_{i,k}^{(1)},\dots,\mathbf{z}_{i,k}^{(M)}\,]$

\State \emph{(a) Pre-trained autoencoder:}
\[
  \mathcal{L}_{AE}
  = \frac1{3N}\sum_{i,k}\frac1{M\,2E}\bigl\|\hat F_{i,k}-F_{i,k}\bigr\|_F^2
  \quad\text{(Eq.~\ref{training})}
\]
\State Update $\psi$ via Adam($\eta \prime$) for $E_{\mathrm{AE}}$ epochs

\State \emph{(b) Fine‐tune for classification:}
\For{$k=1,\dots,K$}
  \State $p_{i,k} = \mathrm{Softmax}\bigl(T_\psi(F_{i,k})\bigr)$
\EndFor
\State Fuse
\[
  p_i \;=\;\frac{p_{i,1}\odot\cdots\odot p_{i,K}}
    {\sum\nolimits_c [\,p_{i,1}\odot\cdots\odot p_{i,K}\,]_c}
  \quad\text{(Eq.~\ref{fine tune})}
\]
\State Minimize
\[
  \mathcal{L}_{NLL} = -\frac1N\sum_i\log p_i[y_i]
\]
\State Update $\psi$ via Adam($\eta \prime \prime$) for $E_{\mathrm{CL}}$ epochs

\State \Return $\hat y_i = \arg\max_c p_i[c]$
\end{algorithmic}
\end{algorithm}

\section{Datasets and Implementation}
\label{dataset}
\subsection{PLAsTiCC Dataset}
We use the official PLAsTiCC dataset from the 2018 Kaggle competition, which provides approximately 7848 simulated LSST light curves in six filters (u, g, r, i, z, y) and labels for 14 astrophysical classes (e.g., Type Ia/II supernovae, RR Lyrae). Curve lengths vary from 50 to 350 epochs, and because observations only occur when each field is visible (weather, scheduling, maintenance), roughly 30 \% of per‐band flux measurements are missing on average.

\subsection{SelfRegulationSCP2 Dataset}
The SelfRegulationSCP2 dataset is a collection of EEG recordings from participants learning to control slow cortical potentials. Each record comprises eight scalp channels sampled at 250 Hz. During each session, subjects receive a visual cue and then attempt either to increase (“up”) or decrease (“down”) their brain signal over a 5 s interval, preceded by a 2 s baseline. There are 200 trials per subject—100 “up” and 100 “down”—resulting in a balanced, two‐class (binary) classification task. This clean, well‐labeled dataset is ideal for evaluating and comparing EEG‐based decoding methods.  

\subsection{Bitcoin Price Trend Dataset}
Spanning daily BTC/USD data from 2015–2023, each record includes OHLCV (open, high, low, close, volume) plus 14 technical indicators: RSI-7, RSI-14, CCI-7, CCI-14, SMA-50, EMA-50, SMA-100, EMA-100, MACD, Bollinger Bands, True Range, ATR-7, and ATR-14. Our target is next-day price movement—classified as \emph{positive} (> +1 \%), \emph{negative} (< – 1 \%), or \emph{stable} ($|\%| \leq 1 \%$). We train and tune on data through 2015 to 2022, and evaluate on 2023 (365 days), yielding an approximate class balance of 34 \% positive, 27 \% negative, and 38 \% stable.

\subsection{Hydropower Research Institute (HRI) Dataset}

This proprietary dataset \cite{hri_dataset} comprises continuous sensor recordings from hydropower sites and the associated powerhouses. Although the raw feed is nominally logged every 30 s, individual channels (91 channels) report at irregular intervals (some hourly, others daily or weekly), leaving substantial gaps. Maintenance logs record the exact timestamp of each failure along with one of 14 high-level cause codes (e.g., Main transformer, Shaft packing, Transmission line). We therefore set \(k=3\) views for each event: we extract three contiguous 6 h windows of sensor readings immediately preceding the failure—covering 0–6 h, 6–12 h, and 12–18 h before the event—and assign the corresponding cause code as the label. In total, this yields 390 events (each with 3 windows), which we split chronologically into 80\% for training, 10\% for validation, and 10\% for testing.

\section{Evaluation Metrics}
\label{evaluation metrics}
We report overall accuracy, per‐class precision, recall, and F$_1$‐scores (both micro‐ and macro‐averaged) to quantify classification performance. To illustrate the impact of applying  variational information bottleneck on representation quality, we visualize raw text embeddings and IB‐filtered embeddings using t‐SNE plots, showing improved cluster separation. We also include normalized confusion matrices to highlight class‐wise true and false positive rates. All metrics and visualizations are computed on held‐out test splits for each dataset, ensuring a consistent and robust assessment of our model.

\section{Implementation}
\label{implementation}
\textit{\underline{PLAsTiCC:}} Each light curve is divided into \(M=10\) equal‐duration segments and embedded via the OpenAI Text Embedding (text-embedding-3-small, 1536 dim) model. VIB filter is trained with \(\text{d}=256\), \(\text{epochs}=100\), \(\text{batch\ size}=4\), \(\text{lr}=1\times10^{-4}\), and \(\beta=1\times10^{-4}\). The Transformer classifier uses \(h=128\), \(n_{\mathrm{head}}=32\), \(L=2\), and \(\text{dim}_{\mathrm{ff}}=128\). The Transformer classifier settings are identical (two layers, \(n_{\mathrm{head}}=32\), \(\text{dim}_{\mathrm{ff}}=128\)), with autoencoder and clustering pretraining for 100 and 50 epochs, respectively.

\textit{\underline{SelfRegulationSCP2 :}} Each EEG trial is divided into \(M=24\) equal‐duration segments and embedded via the OpenAI Text Embedding (text-embedding-3-small, 1536-dim) model. VIB filter is trained with \(\text{d}=256\), \(\text{epochs}=100\), \(\text{batch size}=4\), \(\text{lr}=1\times10^{-4}\), and \(\beta=1\times10^{-4}\). The Transformer classifier uses \(h=128\), \(n_{\mathrm{head}}=32\), \(L=2\), and \(\text{dim}_{\mathrm{ff}}=128\), with autoencoder and clustering pretraining for 50 and 50 epochs, respectively.

\textit{\underline{Bitcoin Price Trend:}} We take the most recent 15 days of data per sample, segmented into \(M=5\) non‐overlapping 3-day windows, and embed each window using the OpenAI Text Embedding (small, 1536 dim) model. The VIB uses the same hyperparameters as above. The Transformer classifier employs \(h=128\), \(n_{\mathrm{head}}=16\), \(L=2\), and \(\text{dim}_{\mathrm{ff}}=128\). Autoencoder pretraining runs for 100 epochs and clustering pretraining for 200 epochs.

\textit{\underline{HRI:}} This is a commercial dataset, so we embed using the nomic-embed-text-v1 (765 dim). We extract three consecutive 6‐hour windows immediately preceding each failure $k=3$, each split into \(M=24\) non‐overlapping chunks. The VIB is trained with \(\text{d}=256\), \(\text{epochs}=100\), \(\text{batch\ size}=4\), \(\text{lr}=1\times10^{-4}\), and \(\beta=1\times10^{-4}\). The Transformer classifier uses \(h=128\), \(n_{\mathrm{head}}=32\), \(L=2\), and \(\text{dim}_{\mathrm{ff}}=128\). We pretrain the autoencoder for 100 epochs and the clustering head for 50 epochs.

To evaluate the effectiveness of our proposed ADEPT pipeline on the proprietary HRI dataset—and in the absence of any publicly available benchmark—we instantiate and compare three classification strategies:
\begin{enumerate}
  \item \emph{Feature extraction + Classifier}: We linearly interpolate missing readings onto a uniform 30\,s grid, slide 15\,\text{min} windows over each event, extract over 9{,}000 time- and frequency-domain features per channel via TSFEL, select the top 100 via mutual information, normalize, and classify with same MHA classifier as the ADEPT framework has.
  \item \emph{ADEPT v1.0}: We serialize each 15\,\text{min} segment and embed it offline to a 768\text{-}dim vector via the \texttt{nomic-embed-text-v1} model, then classify directly.
  \item \emph{ADEPT v2.0}: Our full pipeline, where VIB compresses the 768\text{-}dim embeddings before fusion and classification.
\end{enumerate}

\section{Detailed Results on Predicting the Astrophysical Class of Light Curves}
\label{detailed LSST}
\begin{figure}[htbp]
  \centering
  \includegraphics[width=0.8\textwidth]{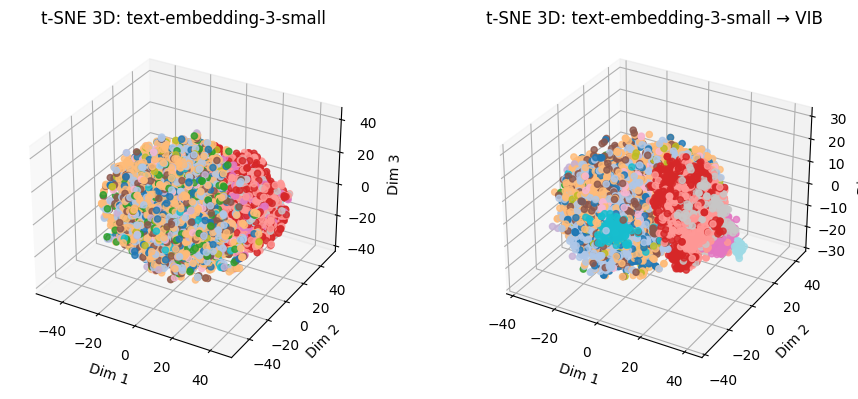}
  \caption{3D t-SNE projection of 1536-dim segment embeddings from the PLAsTiCC-2018 LSST dataset, colored by transient class. \emph{Left:} raw text embeddings; \emph{Right:} embeddings after VIB filtering.}
  \label{fig:LSST_tsne}
\end{figure}

Figure~\ref{fig:LSST_tsne} presents a 3D t-SNE projection of 1536-dimensional segment embeddings from the PLAsTiCC-2018 LSST dataset, colored by transient class (14 astrophysical types). \emph{Left:} raw OpenAI text embeddings exhibit overlapping and diffuse clusters. \emph{Right:} embeddings after Variational Information Bottleneck (VIB) filtering show tighter, well-separated clusters, demonstrating the effectiveness of VIB in ADEPT v2.0.

\begin{figure}[htbp]
    \centering
    \includegraphics[width=0.6\linewidth]{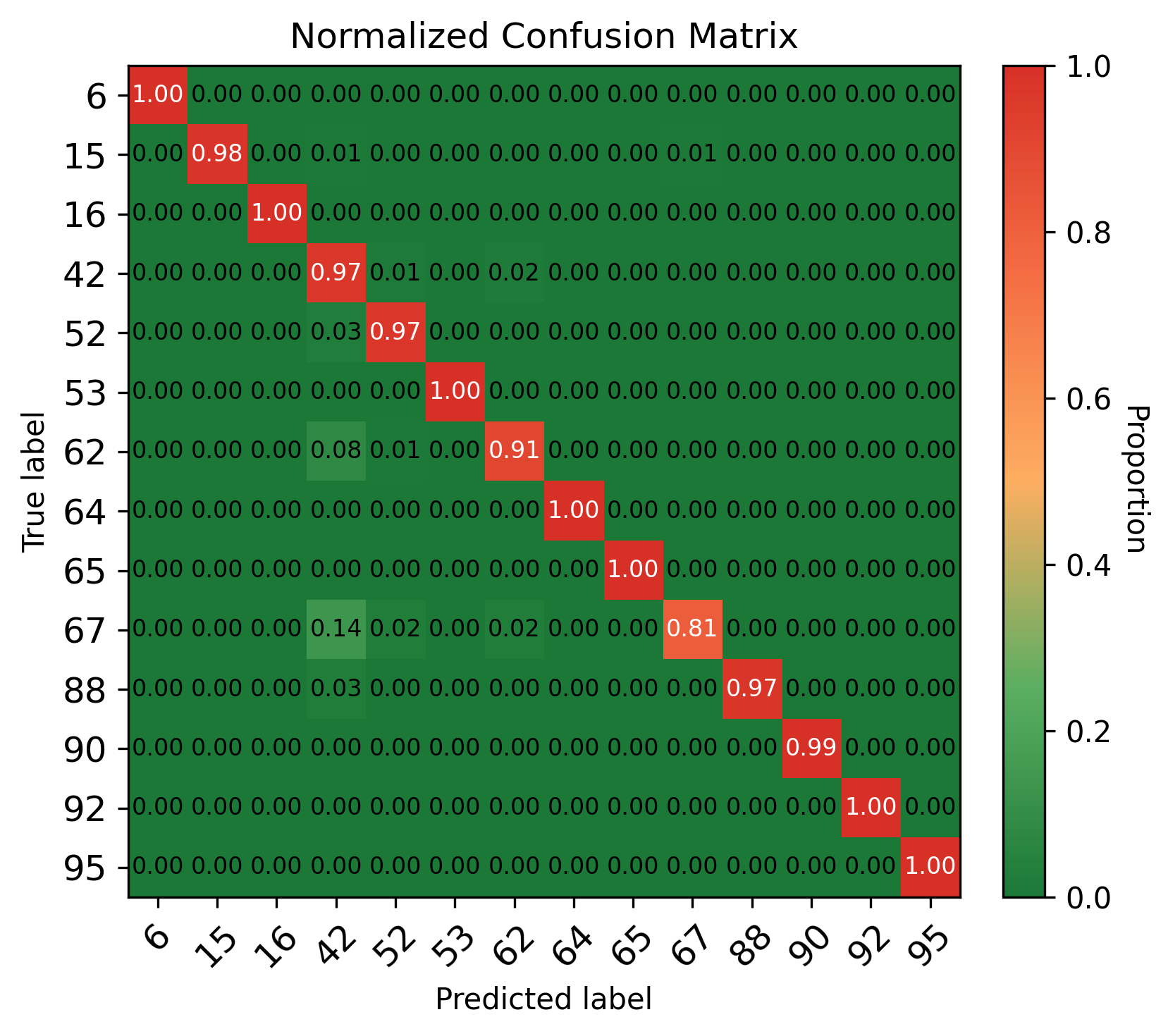}
    \caption{Normalized confusion matrix for the IB‐filtered pipeline on the PLAsTiCC-2018 LSST dataset. Rows correspond to true classes and columns to predicted classes; cell intensity indicates per‐class recall.}
    \label{fig:conf_LSST}
\end{figure}

Table~\ref{tab:lsst_detailed} reports per–class precision, recall, and F$_1$–scores for the 14 astrophysical classes in the PLAsTiCC-2018 test set, alongside overall accuracy (0.98), macro– and weighted–average metrics. Nearly perfect scores are achieved on most classes, with minor drops for classes 42 and 67.
\begin{table}[htbp]
  \centering
  \caption{Per–class performance on PLAsTiCC}
  \label{tab:lsst_detailed}
  \begin{tabular}{lrrrr}
    \toprule
    \textbf{Class} & \textbf{Precision} & \textbf{Recall} & \textbf{F$_1$–score} \\
    \midrule
     6  & 1.00 & 1.00 & 1.00  \\
    15  & 1.00 & 0.98 & 0.99   \\
    16  & 1.00 & 1.00 & 1.00   \\
    42  & 0.92 & 0.97 & 0.94   \\
    52  & 0.85 & 0.97 & 0.91   \\
    53  & 1.00 & 1.00 & 1.00   \\
    62  & 0.94 & 0.91 & 0.92   \\
    64  & 1.00 & 1.00 & 1.00   \\
    65  & 1.00 & 1.00 & 1.00   \\
    67  & 0.94 & 0.81 & 0.87   \\
    88  & 1.00 & 0.97 & 0.99  \\
    90  & 1.00 & 0.99 & 1.00   \\
    92  & 1.00 & 1.00 & 1.00   \\
    95  & 1.00 & 1.00 & 1.00  \\
    \midrule
     \textbf{Accuracy}    & —    & —    & 0.98 \\
    \textbf{Macro avg}    & 0.98 & 0.97 & 0.97 \\
    \textbf{Weighted avg} & 0.98 & 0.98 & 0.98  \\
    \bottomrule
  \end{tabular}
\end{table}
Figure~\ref{fig:conf_LSST} shows the normalized confusion matrix for our VIB embedding pipeline on PLAsTiCC: rows correspond to true classes and columns to predicted classes; cell intensity indicates per‐class recall. Misclassifications are rare and primarily occur among classes with similar light‐curve signatures.

\section{Detailed Results on Predicting Patient Condition
using EEG Data}
\label{detailed EEG}
\begin{figure}[htbp]
\centering
\includegraphics[width=0.8\textwidth]{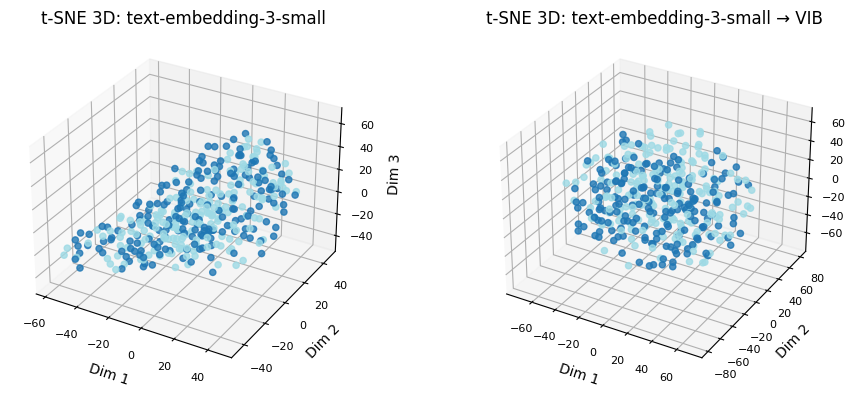}
\caption{3D t‐SNE projection of 1536‐dim segment embeddings for SelfRegulationSCP2 dataset, colored by class. \emph{Left:} raw OpenAI embeddings; \emph{Right:} embeddings after VIB filtering.}
\label{fig:eeg_tsne}
\end{figure}

Figure~\ref{fig:eeg_tsne} presents a 3D t‐SNE projection of 1536‐dimensional segment embeddings for the SelfRegulationSCP2 dataset, colored by class. \emph{Left:} raw OpenAI text‐embedding‐3‐small embeddings exhibit diffuse, overlapping clusters. \emph{Right:} embeddings after Variational Information Bottleneck (VIB) filtering form tighter, more separable clusters.

\noindent Table~\ref{tab:EEG_detailed} reports per–class precision, recall, and F$_1$–scores for the two sentiment classes in our test set, alongside overall accuracy (0.74), macro– and weighted–average metrics. Performance is balanced across classes, with “negativity” achieving 0.76 on all metrics and “positivity” slightly lower at 0.71. Figure~\ref{fig:conf_BTC} shows the normalized confusion matrix for the ADEPT v2.0 pipeline on the SelfRegulationSCP2 dataset. Rows correspond to true movement classes (negativity, positivity) and columns to predicted classes; cell intensity indicates per–class recall.

\begin{table}[htbp]
  \centering
  \caption{Per–class performance on the sentiment classification task.}
  \label{tab:EEG_detailed}
  \begin{tabular}{lrrr}
    \toprule
    \textbf{Class}       & \textbf{Precision} & \textbf{Recall} & \textbf{F$_1$–score}  \\
    \midrule
    Negativity           & 0.76               & 0.76            & 0.76                  \\
    Positivity           & 0.71               & 0.71            & 0.71                  \\
    \midrule
    \textbf{Accuracy}    & —                  & —               & 0.74                  \\
    \textbf{Macro avg.}   & 0.73               & 0.73            & 0.73                  \\
    \textbf{Weighted avg.}& 0.74               & 0.74            & 0.74                  \\
    \bottomrule
  \end{tabular}
\end{table}

\begin{figure}[htbp]
    \centering
    \includegraphics[width=0.4\linewidth]{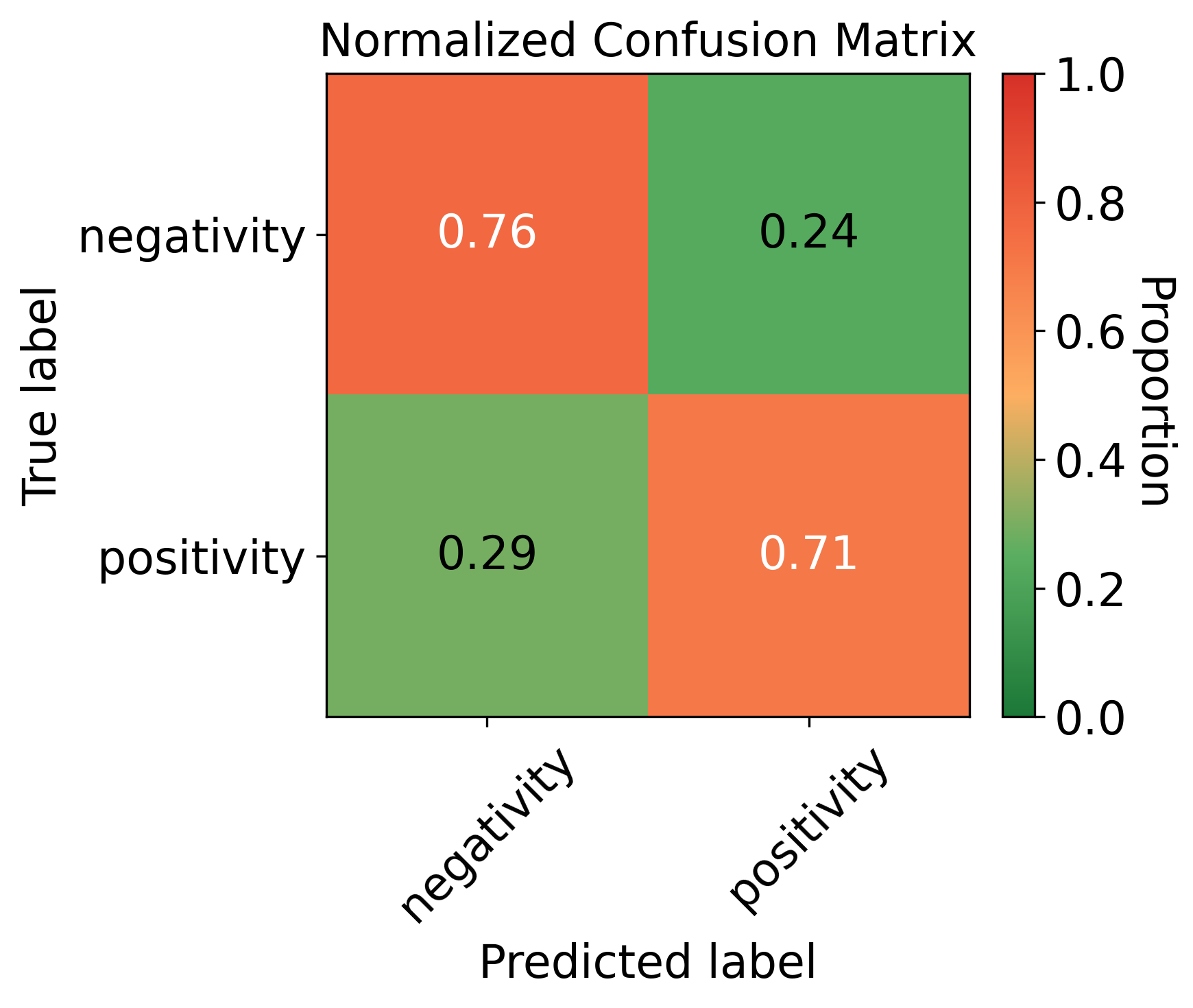}
    \caption{Normalized confusion matrix for the ADEPT v2.0 pipeline on the SelfRegulationSCP2 dataset. Rows correspond to true movement classes (negativity, positivity), columns to predicted classes; cell intensity indicates per–class recall.}
    \label{fig:conf_BTC}
\end{figure}

\section{Detailed Results on Predicting Future Bitcoin Price Trend}
\label{detailed bit}
\begin{figure}[htbp]
  \centering
  \includegraphics[width=0.8\textwidth]{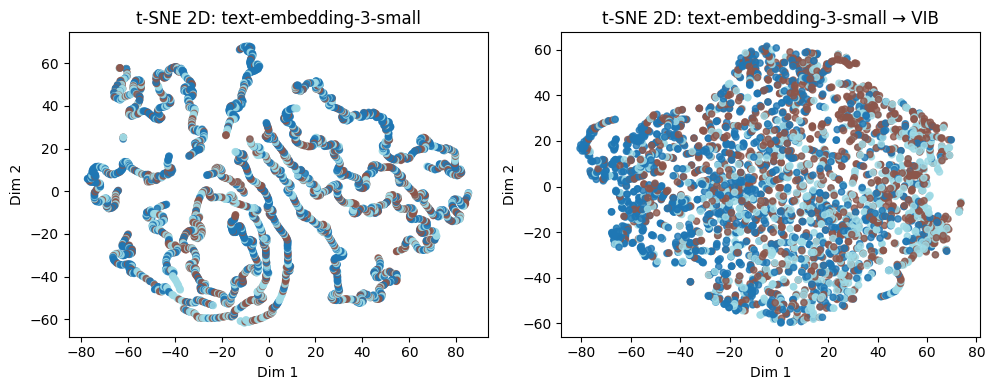}
  \caption{2D t‐SNE projection of text‐serialized Bitcoin segment embeddings (1536 D), colored by next‐day movement class. \emph{Left:} raw OpenAI embeddings; \emph{Right:} embeddings after IB filtering.}
  \label{fig:BTC_tsne}
\end{figure}

Figure~\ref{fig:BTC_tsne} shows 2D t‐SNE visualizations of 1536‐dim segment embeddings from the Bitcoin market dataset, colored by next‐day movement class. \emph{Left:} raw text embeddings form elongated, intertwined trajectories with substantial class overlap. \emph{Right:} VIB filtering embeddings produce more homogeneous clusters for positive, negative, and stable days, indicating enhanced discriminability.

Table~\ref{tab:btc_detailed} reports per‐class precision, recall, and F$_1$‐scores for the three movement classes in the Bitcoin price trend test set, alongside overall accuracy (0.88), macro‐ and weighted‐average metrics. The model achieves strong performance across all classes, with highest F$_1$ on the “stable” class.
\begin{table}[htbp]
  \centering
  \caption{Per‐class performance on the Bitcoin market dataset.}
  \label{tab:btc_detailed}
  \begin{tabular}{lrrr}
    \toprule
    \textbf{Class}      & \textbf{Precision} & \textbf{Recall} & \textbf{F$_1$‐score}  \\
    \midrule
    Long                & 0.87               & 0.88            & 0.87                  \\
    Short               & 0.82               & 0.80            & 0.81                  \\
    Stable              & 0.92               & 0.92            & 0.92                  \\
    \midrule
    \textbf{Accuracy}         & —                  & —               & 0.88                  \\
    \textbf{Macro avg.}        & 0.87               & 0.87            & 0.87                  \\
    \textbf{Weighted avg.}     & 0.88               & 0.88            & 0.88                  \\
    \bottomrule
  \end{tabular}
\end{table}

Figure~\ref{fig:conf_BTC} shows the normalized confusion matrix for our IB‐filtered embedding pipeline on the Bitcoin dataset. Rows correspond to true next‐day movement classes (long, short, stable) and columns to predicted classes; cell intensity indicates per‐class recall.

\begin{figure}[htbp]
    \centering
    \includegraphics[width=0.4\linewidth]{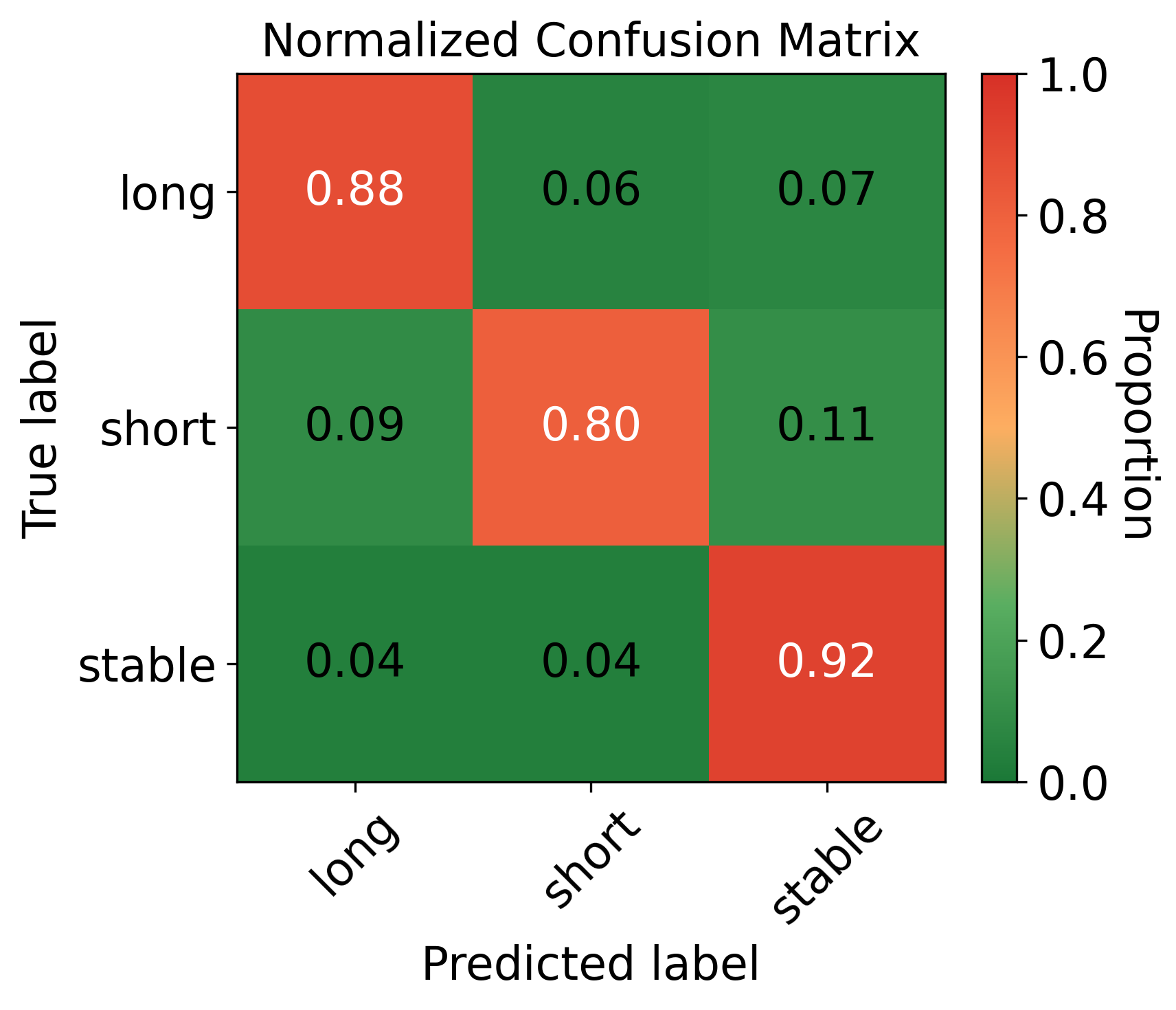}
    \caption{Normalized confusion matrix for the IB‐filtered pipeline on the Bitcoin market dataset. Rows correspond to true movement classes (long, short, stable), columns to predicted classes; cell intensity indicates per‐class recall.}
    \label{fig:conf_BTC}
\end{figure}

\section{Detailed Results on Predicting the Cause of Hydropower Reliability Issues}
\label{detailed HRI}
\begin{figure}[htbp]
  \centering
  \includegraphics[width=0.8\textwidth]{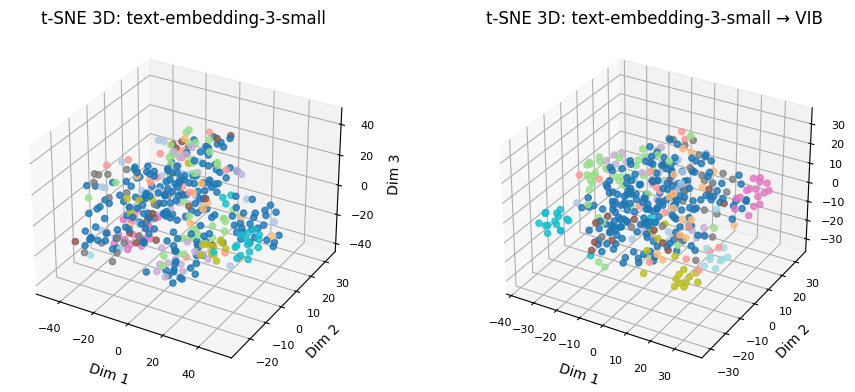}
  \caption{3D t‐SNE projection of 15\,\text{min} segment embeddings from the HRI dataset, colored by failure cause code (13 classes). \emph{Left:} raw 768\text{-}dim Nomic text embeddings; \emph{Right:} embeddings after IB filtering.}
  \label{fig:maintenance_tsne}
\end{figure}

Figure~\ref{fig:maintenance_tsne} shows 3D t‐SNE projections of these 768\text{-}dim segment embeddings colored by cause code: \emph{Left:} raw \texttt{nomic-embed-text-v1} embeddings display diffuse, overlapping clusters; \emph{Right:} IB‐filtered embeddings form compact, well‐separated clusters, indicating improved discriminability and noise suppression.

Table~\ref{tab:hri_detailed} reports per–cause‐code precision, recall, and F$_1$–scores for the 13 failure types in the HRI test set, alongside overall accuracy (0.74), macro– and weighted–average metrics. Perfect F$_1$–scores (1.00) are achieved on several well‐represented classes (e.g., 7030, 7050), while rare classes (e.g., 7009) suffer from zero recall.  
\begin{table}[htbp]
    \centering
    \caption{Per‐class performance on the HRI dataset.}
    \label{tab:hri_detailed}
    \begin{tabular}{lccc}
      \toprule
      \textbf{Class} & \textbf{Precision} & \textbf{Recall} & \textbf{F$_1$–score} \\
      \midrule
      3620 & 1.00 & 0.40 & 0.57 \\
      3710 & 0.50 & 0.50 & 0.50 \\
      4560 & 1.00 & 0.50 & 0.67 \\
      7009 & 0.00 & 0.00 & 0.00 \\
      7030 & 1.00 & 1.00 & 1.00 \\
      7050 & 1.00 & 1.00 & 1.00 \\
      7099 & 0.43 & 1.00 & 0.60 \\
      7110 & 0.89 & 0.76 & 0.82 \\
      9696 & 0.75 & 1.00 & 0.86 \\
      \midrule
      \textbf{Accuracy}    & —    & —    & 0.74 \\
      Macro avg.             & 0.73 & 0.68 & 0.67 \\
      Weighted avg.          & 0.85 & 0.74 & 0.76 \\
      \bottomrule
    \end{tabular}
\end{table}

\begin{figure}[htbp]
    \centering
    \includegraphics[width=0.6\linewidth]{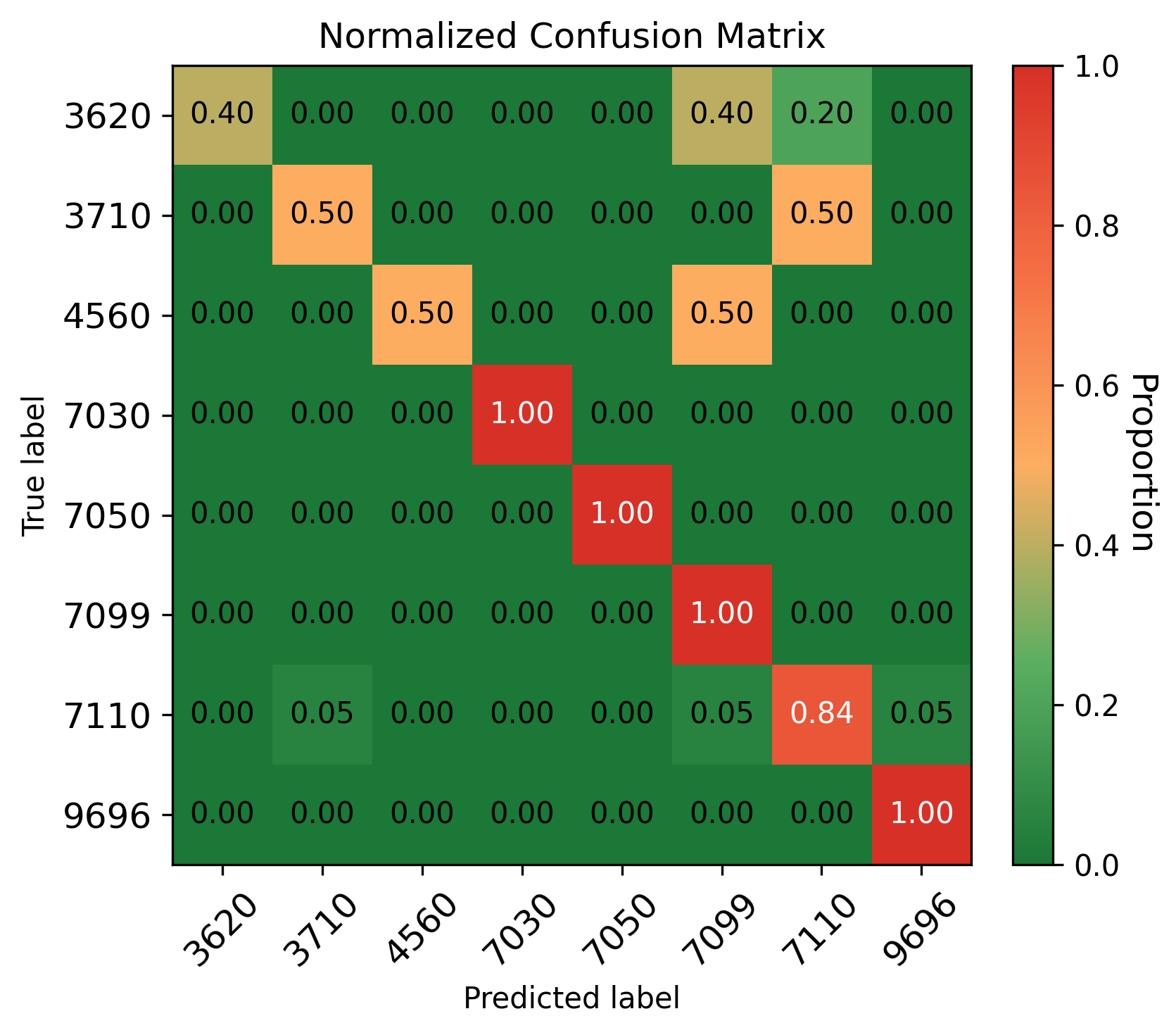}
    \caption{Normalized confusion matrix for the IB‐filtered pipeline on HRI. Rows correspond to true failure codes, columns to predicted codes, and cell intensity indicates per‐class recall.}
    \label{fig:conf_maintenance}
\end{figure}

Figure~\ref{fig:conf_maintenance} shows the normalized confusion matrix for our VIB pipeline. Each row is a true cause code and each column the predicted code; cell intensities indicate per‐class recall. Notable misclassifications occur between codes 3620 and 3710, reflecting similar pre‐failure sensor signatures.

Table \ref{tab:class_probs} reports, for each maintenance event, the model’s Top-1 through Top-3 predicted failure‐mode classes and their associated probabilities, alongside the actual observed class. While the Top-1 selection yields only 74.4 \% accuracy, expanding the recommendation to the Top-2 candidates attains 97.5 \% coverage of the true class. In a maintenance‐industry context—where overlooking the true failure mode can have costly consequences—providing a short ranked list of likely failure modes is therefore far more reliable and actionable than a single “best” guess.

\begin{table}[H]
  \small
  \centering
  \caption{Predicted class probabilities per event (Top-1 and Top-2 only).}
  \label{tab:class_probs}
  \begin{minipage}[t]{0.49\textwidth}
    \vspace{0pt}
    \centering
    \begin{tabularx}{\linewidth}{@{} *{6}{>{\centering\arraybackslash}X}@{}}
      \toprule
      EventID & Actual 
        & \shortstack{Top-1\\Class} 
        & \shortstack{Top-2\\Class} 
        & \shortstack{Top-1\\Prob} 
        & \shortstack{Top-2\\Prob} \\
    \midrule
    6   & 3620 & 7110 & \textbf{3620} & 0.991 & 0.008 \\
    9   & 3620 & \textbf{3620} & 7110 & 0.999 & 0.001 \\
    17  & 3620 & \textbf{3620} & 4600 & 0.851 & 0.143 \\
    39  & 7030 & \textbf{7030} & 4560 & 0.542 & 0.448 \\
    48  & 4560 & \textbf{4560} & 7030 & 0.690 & 0.269 \\
    78  & 4560 & 7099 & \textbf{4560} & 0.999 & 0.001 \\
    85  & 7050 & \textbf{7050} & 4600 & 0.999 & 0.001 \\
    89  & 7050 & \textbf{7050} & 7110 & 0.833 & 0.164\\
    139 & 9696 & \textbf{9696} & 7050 & 0.999 & 0.001 \\
    140 & 9696 & \textbf{9696} & 7050 & 0.999 & 0.001 \\
    143 & 9696 & \textbf{9696} & 7110 & 0.990 & 0.006 \\
    145 & 7110 & 3710 & \textbf{7110} & 0.673 & 0.280 \\
    152 & 7110 & \textbf{7110} & 7050 & 0.995 & 0.002 \\
    163 & 7110 & \textbf{7110} & 3620 & 0.997 & 0.001 \\
    167 & 7110 & \textbf{7110} & 4560 & 0.999 & 0.001 \\
    192 & 7110 & \textbf{7110} & 4560 & 0.999 & 0.001 \\
    202 & 7110 & \textbf{7110} & 3710 & 0.959 & 0.037 \\
    207 & 7099 & \textbf{7099} & 7110 & 0.999 & 0.001 \\
    209 & 7110 & \textbf{7099} & 7110 & 0.935 & 0.057 \\
    215 & 7110 & \textbf{7110} & 7009 & 0.999 & 0.001 \\
    \bottomrule
  \end{tabularx}
\end{minipage}
\hfill
\begin{minipage}[t]{0.49\textwidth}
  \vspace{0pt}
  \centering
  \begin{tabularx}{\linewidth}{@{} *{6}{>{\centering\arraybackslash}X}@{}}
      \toprule
      EventID & Actual 
        & \shortstack{Top-1\\Class} 
        & \shortstack{Top-2\\Class} 
        & \shortstack{Top-1\\Prob} 
        & \shortstack{Top-2\\Prob} \\
      \midrule
    222 & 7110 & \textbf{7110} & 7099 & 0.999 & 0.001 \\
    230 & 3710 & \textbf{3710} & 7099 & 0.999 & 0.000 \\
    233 & 7110 & \textbf{7110} & 7050 & 0.998 & 0.001 \\
    263 & 7110 & \textbf{7110} & 9300 & 0.999 & 0.001 \\
    306 & 7110 & 9696 & \textbf{7110} & 0.523 & 0.404 \\
    347 & 3620 & 7099 & \textbf{3620} & 0.522 & 0.472 \\
    357 & 7110 & \textbf{7110} & 7050 & 0.996 & 0.002 \\
    367 & 7110 & \textbf{7110} & 3620 & 0.999 & 0.001 \\
    402 & 7110 & \textbf{7110} & 9696 & 0.999 & 0.000 \\
    406 & 7110 & \textbf{7110} & 4560 & 0.999 & 0.000 \\
    410 & 7099 & \textbf{7099} & 7009 & 0.999 & 0.0002 \\
    421 & 7110 & \textbf{7110} & 4560 & 0.999 & 0.000 \\
    445 & 7110 & 7009 & \textbf{7110} & 0.821 & 0.144 \\
    478 & 7110 & \textbf{7110} & 9696 & 0.999 & 0.000 \\
    502 & 3620 & 7099 & \textbf{3620} & 0.606 & 0.371 \\
    554 & 7099 & \textbf{7099} & 3710 & 0.999 & 0.000 \\
    560 & 7110 & 7009 & \textbf{7110} & 0.724 & 0.266 \\
    570 & 3710 & 7110 & \textbf{3710} & 0.999 & 0.000 \\
    571 & 7110 & \textbf{7110} & 3710 & 0.879 & 0.116 \\
    \\
    \bottomrule
  \end{tabularx}
\end{minipage}
\end{table}



\end{document}